\pdfoutput=1

\documentclass[11pt]{article}

\usepackage[final]{acl}
\usepackage{listings}

\usepackage{times}
\usepackage{latexsym}
\usepackage{multirow}
\usepackage{colortbl}
\usepackage{xcolor}
\usepackage{amsmath}
\usepackage{amssymb}
\usepackage{enumitem}
\usepackage{mathtools}

\definecolor{cvprblue}{rgb}{0.21,0.49,0.74}
\definecolor{Ocean}{RGB}{129,194,234}
\definecolor{Mygreen}{RGB}{170, 240, 238}

\usepackage[T1]{fontenc}
\usepackage[utf8]{inputenc}
\usepackage{microtype}
\usepackage{inconsolata}
\usepackage{graphicx}

\title{CoRe-MMRAG: Cross-Source Knowledge Reconciliation \\ for Multimodal RAG}

\usepackage{multirow}
\usepackage{colortbl}
\usepackage{xcolor}
\usepackage{booktabs}

\definecolor{cvprblue}{rgb}{0.21,0.49,0.74}
\definecolor{Ocean}{RGB}{129,194,234}
\definecolor{Mygreen}{RGB}{170, 240, 238}

\definecolor{Magenta}{rgb}{0.8, 0.1, 0.6}

\usepackage[utf8]{inputenc}

\usepackage[most]{tcolorbox}

\tcbset{
  promptbox/.style={
    colback=blue!2,            
    colframe=blue!50!black,    
    fonttitle=\bfseries,       
    title=Prompt,              
    boxrule=0.2mm,             
    arc=1mm,                   
    outer arc=1mm,             
    top=1mm,                  
    bottom=1mm,               
    left=1mm,                  
    right=1mm                 
  }
}

\usepackage{listings}
\usepackage{xcolor}
\usepackage{enumitem}

\author{
Yang Tian\footnotemark[3], Fan Liu\footnotemark[2], Jingyuan Zhang\footnotemark[5], Victoria W.\footnotemark[5], Yupeng Hu\footnotemark[3]\thanks{~~Corresponding author}, Liqiang Nie\footnotemark[4]\footnotemark[1] \\
\footnotemark[3] School of Software, Shandong University\\
\footnotemark[2] National University of Singapore, \footnotemark[5] Independent Researcher\\
\footnotemark[4] Harbin Institute of Technology, Shenzhen\\
\small\texttt{\{tianyangchn,liufancs,nieliqiang\}@gmail.com}, \small\texttt{huyupeng@sdu.edu.cn}\\
}

\begin{document}
\maketitle
\begin{abstract}

Multimodal Retrieval-Augmented Generation (MMRAG) has been introduced to enhance Multimodal Large Language Models by incorporating externally retrieved multimodal knowledge, but it introduces two challenges: Parametric-Retrieved Knowledge Inconsistency (PRKI), where discrepancies between parametric and retrieved knowledge create uncertainty in determining reliability, and Visual-Textual Knowledge Inconsistency (VTKI), where misalignment between visual and textual sources disrupts entity representation. To address these challenges, we propose \textbf{C}r\textbf{o}ss-source knowledge \textbf{Re}conciliation for \textbf{M}ulti\textbf{M}odal \textbf{RAG} (CoRe-MMRAG), a novel end-to-end framework that effectively reconciles inconsistencies across knowledge sources. CoRe-MMRAG follows a four-stage pipeline: it first generates an internal response from parametric knowledge, then selects the most relevant multimodal evidence via joint similarity assessment, generates an external response, and finally integrates both to produce a reliable answer. Additionally, a specialized training paradigm enhances knowledge source discrimination, multimodal integration, and unified answer generation. Experiments on KB-VQA benchmarks show that CoRe-MMRAG achieves substantial improvements over baseline methods, achieving 5.6\% and 9.3\% performance gains on InfoSeek and Encyclopedic-VQA, respectively.  We release code and data at \href{https://github.com/TyangJN/CoRe-MMRAG}{https://github.com/TyangJN/CoRe-MMRAG}.
\end{abstract}

\section{Introduction}
Recent advances in Multimodal Large Language Models (MLLMs)~\cite{alayrac2022flamingo, li2023blip, liu2024visual, liu2024improved, achiam2023gpt, reid2024gemini} have significantly improved multimodal reasoning and generation tasks by leveraging joint vision-language representations.
However, these models inherently suffer from hallucination~\cite{bai2024hallucination} and knowledge limitations~\cite{caffagni2024wiki}, as their parametric knowledge is frozen after pretraining and cannot dynamically adapt to external information.

\begin{figure}[t]
  \includegraphics[width=\columnwidth]{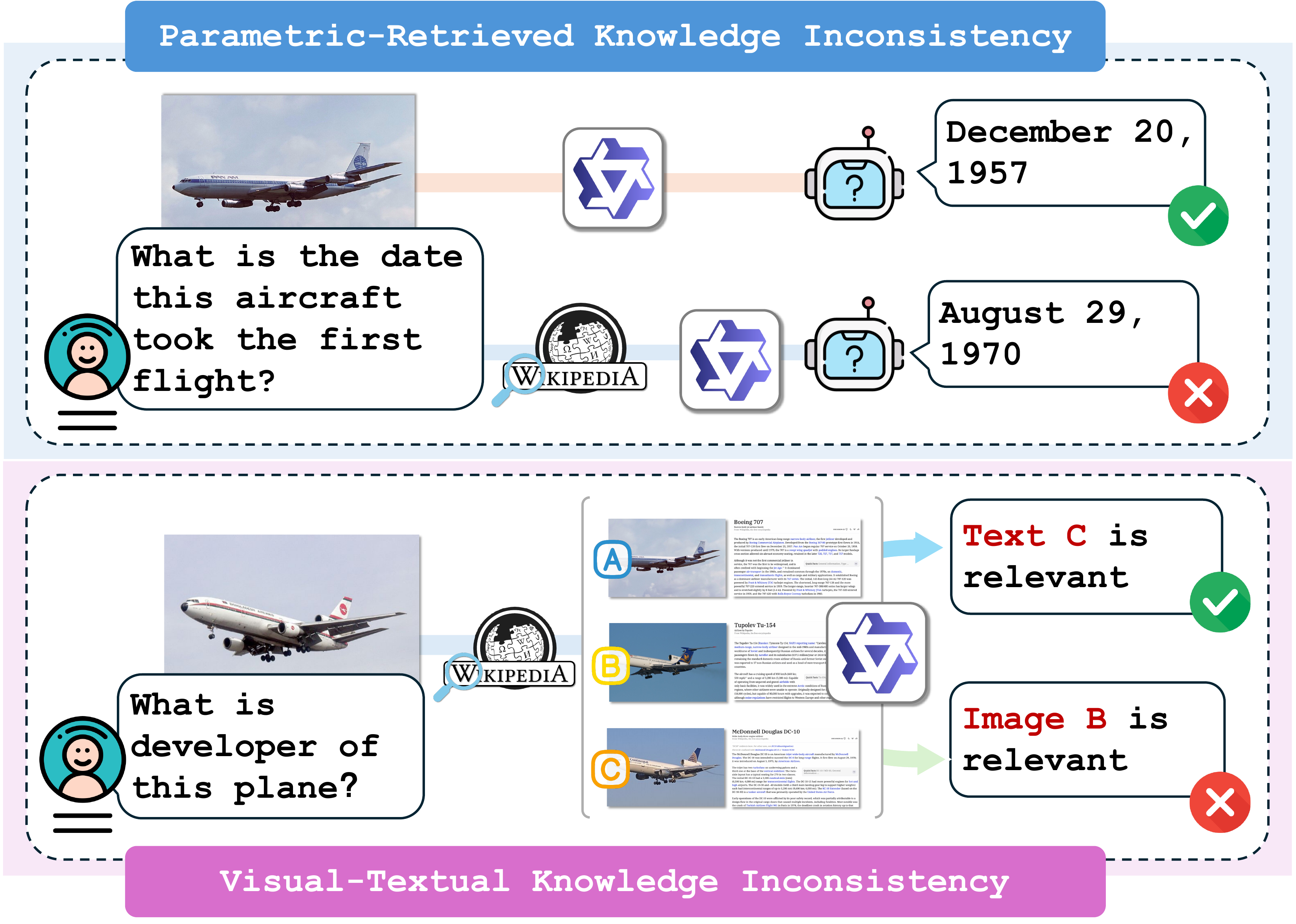}
  \caption{Two types of knowledge inconsistency in MMRAG: (1) Parametric-Retrieved Knowledge Inconsistency, where parametric and retrieved external knowledge generate conflicting answers to the same query. (2) Visual-Textual Knowledge Inconsistency, where misalignments between visual and textual sources disrupt entity representation. }
  \label{fig:motivation}
  \vskip -0.2in
\end{figure}

Multimodal Retrieval-Augmented Generation (MMRAG) has emerged as a promising approach to enhance MLLMs by incorporating retrieved textual and visual knowledge during inference~\cite{yan2024echosight, qi2024rora, zhang2024mr}. By accessing external information, MMRAG helps mitigate knowledge gaps and improves factual grounding. 
However, integrating retrieved knowledge into MLLMs presents two key challenges. 
First, \textbf{Parametric-Retrieved Knowledge Inconsistency (PRKI)}. Since MLLMs rely on frozen pretraining knowledge, retrieved text and images may contradict, extend, or refine this information. Moreover, retrieved content can be incomplete, noisy, or misleading, introducing biases or factual errors. Without effective reconciliation, the model may struggle to balance reliability between internal parametric and external retrieved knowledge, leading to incorrect responses. As shown in Figure~\ref{fig:motivation}, for the aircraft's first flight date, introducing noisy retrieval information (August 29, 1970) overrides the model's reliable parametric knowledge (December 20, 1957). Second, \textbf{Visual-Textual Knowledge Inconsistency (VTKI)}. Since each modality captures different aspects of entity representation~\cite{li2023blip}, retrieved images and textual documents often provide non-overlapping and misaligned information. For instance, an image may visually relate to the query while its paired text describes a different aspect or interpretation. These inconsistencies disrupt knowledge integration, making it difficult for the model to determine which information to prioritize. 

To address the above-mentioned challenges, we propose \textbf{C}r\textbf{o}ss-source knowledge \textbf{Re}conciliation for \textbf{M}ulti\textbf{M}odal RAG (CoRe-MMRAG), a novel end-to-end framework designed to mitigate the inconsistencies between different knowledge sources. CoRe-MMRAG follows a four-stage pipeline: it first generates an initial response using only the model’s internal knowledge, then performs joint similarity assessment to select the most relevant multimodal evidence, followed by generating a response grounded in the retrieved content, and finally integrates both sources to produce a coherent and reliable answer. This structured process enables the model to effectively reconcile diverse knowledge inputs and leverage complementary information from multiple modalities.

To further enhance knowledge reconciliation, CoRe-MMRAG incorporates a specialized training paradigm with three key objectives: knowledge source selection that learns to identify the most reliable information source between internal parametric and external retrieved knowledge, multimodal selection that optimizes the joint understanding of visual-textual pairs, and unified answer generation that ensures consistent and accurate responses. Through this comprehensive training strategy, CoRe-MMRAG develops robust capabilities in handling knowledge inconsistencies between different sources.

We conduct comprehensive experiments on two knowledge-based VQA benchmarks \cite{mensink2023encyclopedic, chen2023can} to evaluate our approach. Using Qwen2-VL-7B \cite{wang2024qwen2} as the base MLLM, CoRe-MMRAG demonstrates substantial improvements in both zero-shot and fine-tuned setting. Specifically, our method achieves performance gains of 5.6\% over the baseline on the InfoSeek, while also surpassing the baseline on the Encyclopedic-VQA benchmark by 9.3\%. The contributions of this work are summarized as follows:
\begin{itemize}
    \item We identify and formalize two fundamental challenges in multimodal RAG: Parametric-Retrieved Knowledge Inconsistency and Visual-Textual Knowledge Inconsistency. To address these issues, we propose CoRe-MMRAG, a novel end-to-end framework that effectively reconciles inconsistencies across different knowledge sources. 
    \item We design a specialized training paradigm with three targeted objectives that enhance MLLMs in knowledge source discrimination, multimodal integration, and unified answer generation, enabling effective knowledge inconsistency mitigation.
    \item Extensive experiments demonstrate that CoRe-MMRAG achieves significant performance gains over previous SOTAs on multiple Knowledge-based VQA benchmarks.
\end{itemize}

\section{Related Work}
\textbf{Multimodal RAG.} 
Recent advancements in MLLMs, such as LLaVA-family \cite{liu2024visual, li2024llava}, Qwen2-VL \cite{wang2024qwen2}, MiniCPM-V \cite{yao2024minicpm}, and Intern-VL \cite{chen2024internvl}, have demonstrated remarkable performance across various multimodal tasks. However, these models inherently suffer from hallucination issues~\cite{li2023evaluating, caffagni2024wiki}. One effective approach to mitigating hallucination is the incorporation of multimodal data \cite{liu2024survey}, which provides complementary knowledge beyond text-based sources. By integrating multimodal information, models can ground their responses in more diverse and concrete evidence, reducing the likelihood of hallucination. Another approach is inspired by RAG\cite{kandpal2023large, li2025encoder, gao2023retrieval, asai2024self}, several multimodal RAG frameworks\cite{lin2023fine, humrag, adjali2024multi} have been proposed, including Wiki-LLaVA \cite{caffagni2024wiki}, EchoSight \cite{yan2024echosight}, RoRA-VLM \cite{qi2024rora} and LLaVA-m$R^2$AG \cite{zhang2024mr}. These methods typically follow a three-stage pipeline of retrieval, reranking, and generation. However, during reranking, they primarily rely on text-based similarity measures between retrieved passages and questions, which may lead to incorrect passage selection due to the inherent cross-modal discrepancy, ultimately affecting the final answer quality.

\textbf{Multimodal Understanding.}
Multimodal understanding \cite{zhang2024vision, yin2024lamm, wang2025video, liu2018attentive} aims to integrate and interpret information across multiple modalities, such as vision and language, to enhance reasoning and decision-making. A key challenge in this field is incorporating external knowledge to support tasks that require information beyond what is directly present in one modality. One major research direction is knowledge-based Visual Question Answering (VQA), where models answer questions requiring external factual knowledge. Early benchmarks like OK-VQA \cite{marino2019ok} and A-OKVQA \cite{schwenk2022okvqa} introduced commonsense and general knowledge reasoning, while ViQuAE \cite{lerner2022viquae} expanded to entity-specific queries. More recent datasets, such as InfoSeek \cite{chen2023can} and Encyclopedic VQA \cite{mensink2023encyclopedic}, enforce both visual grounding and knowledge retrieval, exposing the limitations of existing models, including LLM-based approaches, in handling multimodal knowledge integration. Beyond VQA, multimodal understanding extends to tasks such as image-text retrieval, captioning, and reasoning, where aligning visual and textual representations is critical. 

\section{Methodology}
\begin{figure*}[ht]
    \begin{center}
    \centerline{\includegraphics[width=1.0\textwidth]{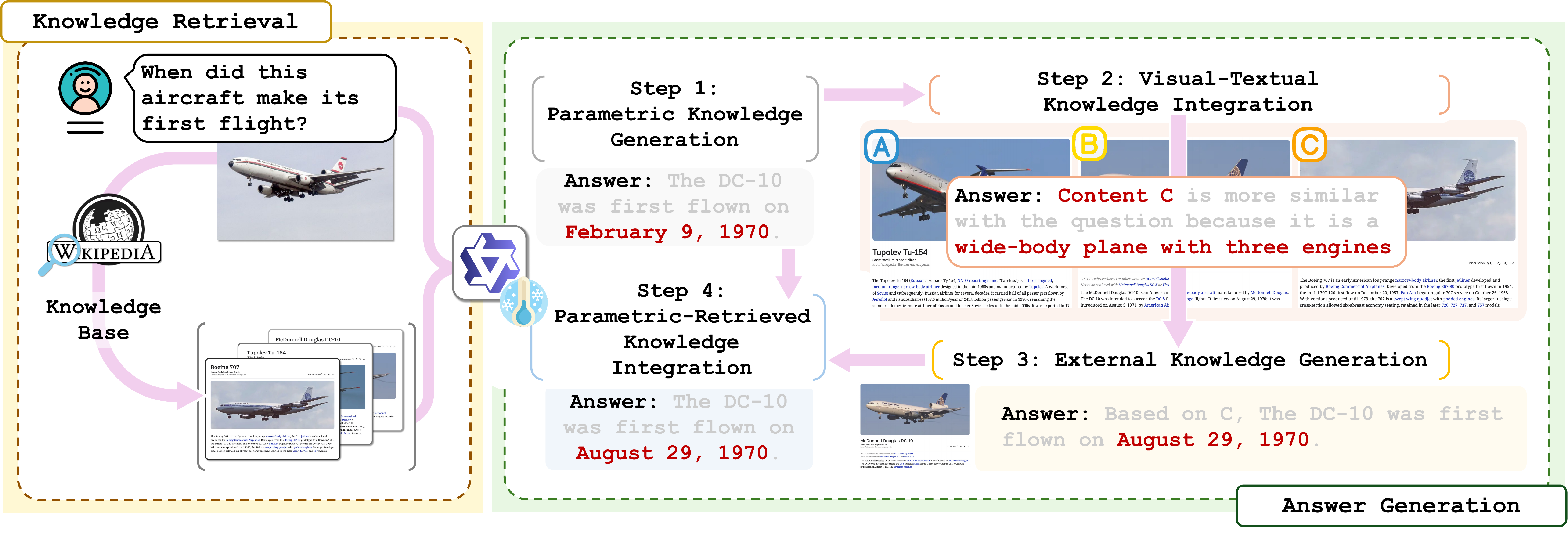}}
    \caption{Overview of the CoRe-MMRAG Framework. CoRe-MMRAG processes a multimodal query and retrieved knowledge in four stages: (1) generating an initial response based solely on parametric knowledge; (2) jointly integrating visual and textual information to identify the most relevant retrieved content; (3) generating an external response based on the selected content from step (2); and (4) reconciling discrepancies between parametric and retrieved knowledge to produce the final answer.}
    \label{fig:framwork}
    \end{center}
\vskip -0.4in
\end{figure*}
In this section, we first formalize the problem of Multimodal Retrieval-Augmented Generation (MMRAG) and introduce two types of inconsistency (\S\ref{sec:problem_formulation}) that arise when applying MMRAG to KB-VQA. We then present our framework (\S\ref{sec:E2E}) and detail our training approach (\S\ref{sec:knowledge_selection}), both of which are designed to enhance the capacity of MLLMs to resolve knowledge inconsistency.

\subsection{Problem Formulation}
\label{sec:problem_formulation}
To assess the effectiveness of our proposed framework, we conduct evaluations on Knowledge-based Visual Question Answering (KB-VQA) tasks. Given an input image-question pair $Q=(Q^v, Q^t)$ from the question set $\textbf{\textrm{Q}}$ with support from an external knowledge base $\textbf{\textrm{K}}$, where each knowledge entry $\textbf{\textrm{K}}_i$ comprises a visual component $V_i$ and the associated textual article $T_i$. A typical MMRAG pipeline addresses KB-VQA through three stages: retrieval, reranking, and generation. In the retrieval stage, a frozen CLIP \cite{radford2021learning} encodes both the query image $Q^v$ and knowledge base images $\{V_i\}_{i=1}^{|\textbf{\textrm{K}}|}$ into a shared embedding space, where the relevance is measured via cosine similarity: $\mathrm{sim}(Q^v, V_i)$. The top-$k$ entries $\{(V_i, T_i)\}_{i=1}^{k}$ are retrieved based on these similarity scores. Then, the retrieved entries are reranked by the multimodal model $\mathcal{M}$ based on the semantic relevance between $Q$ and the retrieved articles $\{T_i\}_{i=1}^{k}$. Finally, $\mathcal{M}$ processes $Q$ along with the most relevant entry to generate the final answer.

During this process, we identify two types of issues. The first problem arises from the inconsistency between the model's parametric knowledge and the retrieved external knowledge, referred to as \textbf{Parametric-Retrieved Knowledge Inconsistency (PRKI)}. Formally, let $\mathcal{M}(Q)$ denote the output based on the model's parametric knowledge, and let $\mathcal{M}(Q, \textbf{\textrm{P}})$ represent the response when augmented with the retrieved knowledge set $\textbf{\textrm{P}}=\{(V_i, T_i)\}_{i=1}^{k}$. The PRKI occurs when:
\begin{equation}
    \mathcal{M}(Q) \neq \mathcal{M}(Q, \textbf{\textrm{P}}). 
\end{equation}
The second issue is \textbf{Visual-Textual Knowledge Inconsistency (VTKI)}, which arises when textual and visual modalities of the retrieved entries $\{(T_i,V_i)\}_{i=1}^k$ yield inconsistent relevance rankings. Formally, the VTKI manifests when:
\begin{equation}
    \operatorname{argmax}r^{\mathcal{M}}(Q,\textbf{\textrm{T}}) \ne \operatorname{argmax}r^{\mathcal{M}}(Q,\textbf{\textrm{V}}),
\end{equation}
where $\operatorname{argmax} r^{\mathcal{M}}(\cdot,\cdot)$ represents model $\mathcal{M}$'s prediction of the most relevant entry ID in each modality. Both PRKI and VTKI can significantly impact model performance. PRKI occurs when noisy external knowledge overrides reliable parametric knowledge, resulting in erroneous outputs. Meanwhile, VTKI becomes especially problematic during the reranking stage. Due to the inconsistency between the textual and visual knowledge, relying on unimodal knowledge increases the risk of introducing irrelevant information to $\mathcal{M}$, which can propagate errors from the reranking stage to answer generation, potentially leading to PRKI and a reduction in model performance.

\subsection{\texorpdfstring{CoRe-MMRAG Framework}{CoRe-MMRAG Framework}}
\label{sec:E2E}

To mitigate the PRKI and VTKI problems described in~\S\ref{sec:problem_formulation}, we propose CoRe-MMRAG, a framework that effectively reconciles inconsistencies across knowledge sources. As shown in Figure~\ref{fig:framwork}, given a query $Q$ and its retrieved knowledge entries $\textbf{\textrm{P}}$, the model $\mathcal{M}$ is prompted to generate responses across four distinct stages in an end-to-end manner. Below, we detail each stage of the framework.

\textbf{Step 1: Parametric Knowledge Generation}. Although external knowledge $\textbf{\textrm{P}}$ is available in the input, we first prompt the model to generate $y^{\mathrm{int}}$ based solely on its parametric knowledge:

\begin{equation} 
    y^{\mathrm{int}} = \mathcal{M}(Q).
\end{equation}

This generation establishes a reference point for detecting potential conflicts with retrieved knowledge in Step 4.

\textbf{Step 2: Visual-Textual Knowledge Integration}. Following the parametric knowledge generation, the model evaluates the relevance between query $Q$ and knowledge entries in $\textbf{\textrm{P}}$. Considering the potential VTKI manifested as:
\begin{equation}
    \left\{
    \begin{aligned}
    I^v & =\underset{i \in \{1, \dots, k\}}{\operatorname{argmax}}\ r^{\mathcal{M}}(Q, \{V_i\}_{i=1}^k),  \\
    I^t & = \underset{j \in \{1, \dots, k\}}{\operatorname{argmax}}\ r^{\mathcal{M}}(Q, \{T_j\}_{j=1}^k), \\
    I^t & \ne I^v.
    \end{aligned}
\right.
\end{equation}
We propose a joint similarity assessment that utilizes the complementary nature of visual and textual modalities:
\begin{equation}
    I^{tv}  = \underset{i \in \{1, \dots, k\}}{\operatorname{argmax}} r^{\mathcal{M}}(Q, \{V_i, T_i\}_{i=1}^k),  \\
\end{equation}
where $I^{tv}$ denotes the most relevant entry ID based on multimodal knowledge. This unified ranking approach jointly leverages abstract semantic concepts from textual descriptions and detailed visual characteristics, eliminating the bias introduced by separate unimodal evaluations, which enables a more robust relevance assessment and resolves VTKI.

\textbf{Step 3: External Knowledge Generation.} After obtaining the most relevant knowledge entry $\textbf{\textrm{P}}_{I^{tv}}$, the model is prompted to generate a response based on the retrieved external knowledge:
\begin{equation}
    y^{\mathrm{ext}} = \mathcal{M}\big(Q, (V_{I^{tv}}, T_{I^{tv}})\big),
\end{equation}
where $y^{\mathrm{ext}}$ denotes the model's response that explicitly considers the retrieved visual-textual knowledge pair $(V_{I^{tv}}, T_{I^{tv}})$. 

\textbf{Step 4: Parametric-Retrieved Knowledge Integration.} 
Given responses from parametric knowledge $y^{\mathrm{int}}$ and retrieved knowledge $y^{\mathrm{ext}}$, the parametric-retrieved knowledge inconsistencies may arise. The model is prompted to resolve these inconsistencies and generate the final response:
\begin{equation}
    y^* = \mathcal{M}\big(Q, y^{\mathrm{int}}, y^{\mathrm{ext}}, (V_{I^{tv}}, T_{I^{tv}})\big),
\end{equation}
where $ y*$ represents the final response, which is determined by comparing the credibility of parametric knowledge and retrieved external knowledge. This process enables the model to leverage both knowledge sources while ensuring the reliable generation of the final answer.

\subsection{Inconsistency-Aware Multimodal Training}
\label{sec:knowledge_selection} 
Inspired by the self-training mechanism in STaR \cite{zelikman2022star}, we propose a fine-tuning paradigm that leverages the model's self-generated outputs under different knowledge conditions. The model learns to resolve PRKI and mitigate VTKI through three specialized training objectives, thus improving its ability to generate accurate answers based on retrieved knowledge.

\textbf{Parametric-Retrieved Knowledge Selection}. We begin by generating answers based solely on the model’s parametric knowledge, $\hat{y}^{\mathrm{int}}=\mathcal{M}(Q)$, and re-evaluate the same question with retrieved external knowledge, obtaining $\hat{y}^{\mathrm{ext}}=\mathcal{M}(Q,\textbf{\textrm{P}})$. After generating both outputs, we filter questions where the model produces correct answers exclusively using either internal or external knowledge, forming fine-tuning datasets $\textbf{\textrm{D}}^{\mathrm{int}}$ and $\textbf{\textrm{D}}^{\mathrm{ext}}$:  
\begin{equation}
    \left\{
    \begin{aligned}
    &\hat{y}^{\mathrm{int}}\ne\hat{y}^{\mathrm{ext}},\\
    &\textbf{\textrm{D}}^{\mathrm{int}}=\{(Q_j,\textbf{\textrm{P}}_j)|\hat{y}_j^{\mathrm{in t}}=\hat{y}^{\mathrm{gt}}_j\}_{j=1}^{|\textbf{\textrm{Q}}|},\\
    &\textbf{\textrm{D}}^{\mathrm{ext}}=\{(Q_i,\textbf{\textrm{P}}_i)|\hat{y}_i^{\mathrm{ext}}=\hat{y}^{\mathrm{gt}}_i\}_{i=1}^{|\textbf{\textrm{Q}}|}.\\
    \end{aligned}
\right.
\end{equation}
Then, we fine-tune the model $\mathcal{M}$ on $\textbf{\textrm{D}}^{\mathrm{int}}$ and $\textbf{\textrm{D}}^{\mathrm{ext}}$, the training is guided by loss function $\mathcal{L}_{\mathrm{PRKI}}$:
\begin{equation}
   \begin{aligned}
        \mathcal{L}_{\mathrm{PRKI}} = - \bigg[
        &\sum_{\mathclap{(Q_j, \textbf{\textrm{P}}_j)\sim \textbf{\textrm{D}}^{\mathrm{int}}}}\ \log \mathcal{M}(\hat{y}^{\mathrm{int}}|Q_j, \textbf{\textrm{P}}_j) + \\
        &\sum_{\mathclap{(Q_i, \textbf{\textrm{P}}_i)\sim \textbf{\textrm{D}}^{\mathrm{ext}}}}\ \log \mathcal{M}(\hat{y}^{\mathrm{e x t}}|Q_i, \textbf{\textrm{P}}_i)\bigg],
    \end{aligned} 
    \label{eq:9}
\end{equation} 
which encourages the model to prioritize the knowledge source that leads to correct answer generation, ensuring robustness in handling PRKI.

\textbf{Visual-Textual Knowledge Selection}. The model computes the most relevant entry IDs independently using visual knowledge $\hat{I}^{v}= \operatorname{argmax}r^{\mathcal{M}}(Q,\textbf{\textrm{V}})$ and textual knowledge $\hat{I}^{t}= \operatorname{argmax}r^{\mathcal{M}}(Q,\textbf{\textrm{T}})$. The training datasets $\textbf{\textrm{D}}^{v}$ and $\textbf{\textrm{D}}^{t}$ are constructed by selecting samples as follows:
\begin{equation}
    \left\{
    \begin{aligned}
    &\hat{I}^{v}\ne\hat{I}^{t},\\
    &\textbf{\textrm{D}}^{v}=\{(Q_m,\textbf{\textrm{P}}_m)|\hat{I}_m^{v}=I^{\mathrm{gt}}_m\}_{m=1}^{|\textbf{\textrm{Q}}|},\\
    &\textbf{\textrm{D}}^{t}=\{(Q_n,\textbf{\textrm{P}}_n)|\hat{I}_n^{t}=I^{\mathrm{gt}}_n\}_{n=1}^{|\textbf{\textrm{Q}}|}.\\
    \end{aligned}
\right.
\end{equation}
Here, $I^{\mathrm{gt}}$ denotes the ground-truth index for the most relevant entry. The model is then fine-tuned on $\textbf{\textrm{D}}^{v}$ and $\textbf{\textrm{D}}^{t}$ using:
\begin{equation}
   \begin{aligned}
        \mathcal{L}_{\mathrm{VTKI}} = - \bigg[
        &\sum_{\mathclap{(Q_m, \textbf{\textrm{P}}_m)\sim \textbf{\textrm{D}}^{v}}}\ \log \underset{j \in \{1, \dots, k\}}{\arg\max}\ r^\mathcal{M}(\hat{I}^{v}|Q_m, \textbf{\textrm{P}}_m^j) + \\
        &\sum_{\mathclap{(Q_n, \textbf{\textrm{P}}_n)\sim \textbf{\textrm{D}}^{t}}}\ \log \underset{i \in \{1, \dots, k\}}{\arg\max}\ r^\mathcal{M}(\hat{I}^{t}|Q_n, \textbf{\textrm{P}}_n^i)\bigg], 
    \end{aligned} 
    \label{eq:11}
\end{equation} 
where the loss function $\mathcal{L}^{\mathrm{VTKI}}$ enables the model to evaluate the reliability of visual and textual modalities and prioritize the more confident one, thereby mitigating VTKI induced by unimodal bias.

\textbf{Unified Answer Generation}. We countinue training the model on $\textbf{\textrm{D}}^{v}$ and $\textbf{\textrm{D}}^{t}$, applying Supervised Fine-Tuning (SFT) with the loss function:
\begin{equation}
   \begin{aligned}
        \mathcal{L}_{\mathrm{SFT}} = - \bigg[
        &\sum_{\mathclap{(Q_m, \textbf{\textrm{P}}_m)\sim \textbf{\textrm{D}}^{v}}}\ \log \mathcal{M}(\hat{y}^{\mathrm{gt}}|Q_m, \textbf{\textrm{P}}_m^{I^v}) + \\
        &\sum_{\mathclap{(Q_n, \textbf{\textrm{P}}_n)\sim \textbf{\textrm{D}}^{t}}}\ \log \mathcal{M}(\hat{y}^{\mathrm{gt}}|Q_n, \textbf{\textrm{P}}_n^{I^t})\bigg],
    \end{aligned} 
    \label{eq:12}
\end{equation} 
where $\mathcal{L}_{\mathrm{SFT}}$ is used to fine-tune the model, encouraging it to generate accurate answers based on the ground-truth external knowledge.

\section{Experiments}
\subsection{Datasets}
\begin{table}[t]
    \begin{center}
    \begin{small}
        \begin{tabular}{lcccc}
        \toprule
        Dataset & Train & Val & Test & KB \\
        \midrule
        Enc-VQA & 1M & 13.6K & 5.8K & 2M Wiki \\
        InfoSeek & 934K & 73K & 348K & 100K Wiki \\ 
        \bottomrule
        \end{tabular}
    \end{small}
    \end{center}
    \caption{Statistics of datasets, including sample splits and their corresponding knowledge base sizes.}
    \label{tab: dataset}
    \vskip -0.2in
\end{table}

We evaluate our proposed CoRe-MMRAG on two large-scale knowledge-based VQA benchmarks: Encyclopedic VQA \cite{mensink2023encyclopedic} and InfoSeek \cite{chen2023can}. Both datasets contain diverse visual-textual queries requiring fine-grained entity knowledge, with explicit knowledge bases to ensure answer verifiability. Encyclopedic VQA (Enc-VQA) contains 221K $(Q^t,\hat{y}^{\mathrm{gt}})$ unique question-answer pairs distributed across 16.7K fine-grained entities, where each question-answer pair is associated with up to five diverse instance images, resulting in 1M image-question-answer $(Q^v,Q^t,\hat{y}^{\mathrm{gt}})$ triplets, while InfoSeek contains 1.3M $(Q^v,Q^t,\hat{y}^{\mathrm{gt}})$ triplets corresponding to approximately 11K distinct entities. Detailed statistics for both data sets, including sample splits and knowledge base sizes, are shown in Table~\ref{tab: dataset}. Following standard practice in EchoSight \cite{yan2024echosight}, we evaluate on Enc-VQA's test set after excluding two-hop questions, resulting in 4.7K test triplets. For InfoSeek, we report results on the validation split, which contains unseen entities (Unseen-E) and novel questions (Unseen-Q).

\subsection{Metrics}
\textbf{Metrics for Retrieval.} We adopt Recall@$k$ as the standard metric to evaluate the retrieval performance~\cite{Liu2021IMPGCN}. This metric examines whether the ground-truth article appears within the top-$k$ retrieved results. Following EchoSight, the evaluation criterion considers an article correct only when its URL exactly matches the ground-truth URL, ensuring precise assessment of retrieval accuracy.

\textbf{Metrics for Answer Generation.} We employ dataset-specific metrics following standard practices in knowledge-based VQA. For Enc-VQA, we use BEM \cite{zhang2019bertscore}, while for InfoSeek \cite{chen2023can}, we adopt both VQA accuracy \cite{marino2019ok} and Relaxed Accuracy \cite{methani2020plotqa, masry2022chartqa}.

\subsection{Implementation Details}
\textbf{Retriever.} For external knowledge retrieval, Enc-VQA utilizes a knowledge base consisting of 2M Wikipedia articles and up to 5M associated images. In contrast, InfoSeek employs a filtered subset of 100K Wikipedia articles with approximately 350K images, following the setup in~\cite{yan2024echosight}. Visual features are extracted using a frozen Eva-CLIP-8B encoder \cite{sun2023eva}, where we use the pooled embeddings from the last layer to compute the cosine similarity between reference images and candidate images. We construct a visual feature index using the FAISS library for efficient similarity search and retrieve the top-5 most relevant entries as external knowledge. Retrieval performance is reported in Table~\ref{tab: retrieval_result}. 

\textbf{Zero-shot Settings.} We employ Qwen2-VL-7B \cite{wang2024qwen2} as our base model, leveraging its 32K token input capacity to accommodate both visual and textual knowledge. To ensure a fair comparison with existing MMRAG approaches, including Wiki-LLaVA \cite{caffagni2024wiki}, EchoSight \cite{yan2024echosight}, RoRA-VLM \cite{qi2024rora}, and LLaVA-mR$^2$AG \cite{zhang2024mr}, we reimplement these pipelines using Qwen2-VL-7B as a unified backbone. We consider the following configurations for zero-shot evaluation: (1)~\textbf{Qwen2-VL-Param}, which relies solely on the model's internal parametric knowledge without any external context; (2)~\textbf{Qwen2-VL-Oracle}, which takes the ground-truth wiki entry as input and serves as an upper-bound; (3) \textbf{Qwen2-VL-1-Stage}, which replicates Wiki-LLaVA's one-stage pipeline by encoding all retrieved entries for answer generation; (4) \textbf{Qwen2-VL-2-Stage}, which follows the two-stage architecture of LLaVA-mR$^2$AG and EchoSight by reranking retrieved entries and generating answers based on the top-ranked one; and (5) \textbf{Qwen2-VL-MMSTaR}, which incorporates all retrieved entries into a Chain-of-Thought reasoning process following the STaR framework~\cite{zelikman2022star}.

\textbf{Fine-tuned Setting}. We fine-tune the Qwen2-VL variants with task-specific supervision to enhance both retrieval accuracy and answer generation. \textbf{Qwen2-VL-1-Stage} is trained using a standard supervised objective $\mathcal{L}_{\mathrm{SFT}}$ with ground-truth wiki entries as input, enhancing the model's ability to generate correct answers from these references. \textbf{Qwen2-VL-2-Stage} is optimized with a combination of a selection objective $\mathcal{L}_{\mathrm{PRKI}}$ and generation objective $\mathcal{L}_{\mathrm{SFT}}$, enabling the model to better identify the correct entry from the top-5 retrieved candidates and generate more accurate answers based on the selected context.

Our proposed CoRe-MMRAG is trained with three objectives, including $\mathcal{L}_{\mathrm{VTKI}}$, $\mathcal{L}_{\mathrm{PRKI}}$, and $\mathcal{L}_{\mathrm{SFT}}$, which jointly enhance parametric-retrieved knowledge selection, visual-textual knowledge selection, and final answer generation. We sample approximately 30K $(Q^v,Q^t,\hat{y}^{\mathrm{gt}})$ triplets from the training set of each benchmark to construct the training set. The model is fine-tuned using LoRA with a rank of 8, applied to all layers. Training is conducted for 3 epochs with a learning rate of $1 \times 10^{-4}$ and a batch size of $1 \times 2$, using 8 H100 GPUs. The full training process takes approximately 10 hours.

\begin{table}[t]
    \vskip -0.0in
    \begin{center}
    \begin{small}
        \begin{tabular}{lcccc}
        \toprule
        \multirow{2}{*}{Datasets} & \multicolumn{4}{c}{Recall@$k$ (\%)} \\
        \cmidrule(r){2-5}
         & $k$=1 & $k$=2 & $k$=5 & $k$=10 \\
        \midrule
        InfoSeek & 45.6 & 54.8 & 67.1 & 73.0 \\
        \midrule
        Enc-VQA & 13.3 & 16.9 & 31.3 & 41.0 \\
        \bottomrule
        \end{tabular}
    \end{small}
    \end{center}
    \caption{The retrieval performance of Eva-CLIP-8B on two datasets.}
    \label{tab: retrieval_result}
    \vskip -0.2in
\end{table}

\subsection{Main Results}
\begin{table*}[t]
   \vskip -0.00in
   \centering
   \setlength{\tabcolsep}{.45em}
   \resizebox{0.85\linewidth}{!}{
       \begin{tabular}{lcccccccccc}
       \toprule
        &  &  &  &  & \textbf{Enc-VQA} &  & \multicolumn{3}{c}{\textbf{InfoSeek}} \\ 
        \cmidrule{6-6} \cmidrule{8-10} 
       \textbf{Model} & \textbf{LLM} &  & \textbf{KB} &  & Single-Hop &  & Unseen-Q & Unseen-E & All \\ 
       \midrule
       \textbf{Zero-shot Models} &  &  &  &  &  &  &  &  &  \\
       \hspace{0.4cm}LLaVA-1.5 & Vicuna-7B &  & - &  & 16.3 &  & 13.0 & 10.3 & 12.2 \\
       \hspace{0.4cm}Qwen2-VL-Param & Qwen2-7B &  & - &  & 12.7 &  & 23.1 & 21.8 & 22.1 \\
       \hspace{0.4cm}Qwen2-VL-Oracle & Qwen2-7B &  & Wiki &  & 51.2 &  & 57.9 & 57.9 & 57.9 \\
       \rowcolor{Ocean!20}
       \hspace{0.4cm}Qwen2-VL-1-Stage$\dagger$ & Qwen2-7B &  & Wiki &  & 17.9 &  & 40.8 & 40.9 & 40.9 \\
       \rowcolor{Ocean!20}
       \hspace{0.4cm}Qwen2-VL-2-Stage$\dagger$ & Qwen2-7B &  & Wiki &  &17.0 &  & 34.9 & 35.1 & 35.0 \\
       \rowcolor{Ocean!20}
       \hspace{0.4cm}Qwen2-VL-MMSTaR$\dagger$ & Qwen2-7B &  & Wiki &  & 16.9 &  & 33.4 & 34.0 & 33.9 \\
       \rowcolor{Mygreen!20}
       \hspace{0.4cm}\textbf{Ours}$\dagger$& Qwen2-7B &  & Wiki &  & \textbf{20.1} &  & \textbf{42.3} & \textbf{43.3} & \textbf{42.9} \\ 
       \midrule
       \textbf{Fine-tuned Models} &  &  &  &  &  &  &  &  &  \\
       \hspace{0.4cm}Wiki-LLaVA & Vicuna-7B &  & Wiki &  & 17.7 &  & 30.1 & 27.8 & 28.9 \\
       \hspace{0.4cm}RoRA-VLM & Vicuna-7B &  & Wiki+Web &  & 20.3 &  & 27.3 & 25.1 & 26.9 \\
       \hspace{0.4cm}EchoSight & LLaMA3-8B &  & Wiki &  & 19.4 &  & - & - & 27.7 \\
       \hspace{0.4cm}LLaVA-mR$^2$AG & Vicuna-7B &  & Wiki &  & 55.1* &  & 39.1 & 39.7 & 39.4 \\
       \rowcolor{Ocean!20}
       \hspace{0.4cm}Qwen2-VL-1-Stage$\dagger$ & Qwen2-7B &  & Wiki &  & 24.3 &  & 42.3 & 43.4 & 43.0 \\
       \rowcolor{Ocean!20}
       \hspace{0.4cm}Qwen2-VL-2-Stage$\dagger$ & Qwen2-7B &  & Wiki &  & 23.1 &  & 36.8 & 37.6 & 37.3 \\
       \rowcolor{Ocean!20}
       \hspace{0.4cm}Qwen2-VL-MMSTaR$\dagger$ & Qwen2-7B &  & Wiki &  & 20.9 &  & 34.8 & 35.2 & 35.1 \\
       \rowcolor{Mygreen!20}
       \hspace{0.4cm}\textbf{Ours}$\dagger$ & Qwen2-7B &  & Wiki &  & \textbf{27.2} &  &\textbf{45.2} & \textbf{46.9} & \textbf{46.5} \\ 
       \bottomrule
       \end{tabular}
   }
   \caption{Main results (\%) on Enc-VQA and InfoSeek with external knowledge. $\dagger$ denotes our method and variants. Note: on Enc-VQA LLaVA-mR$^2$AG* uses Google Lens for retrieval, achieving 62.5\% Recall@5, while our Eva-CLIP-based retrieval achieves 31.3\% Recall@5.}
   \label{tab:main_results}
   \vskip -0.2in
\end{table*}

Table~\ref{tab:main_results} presents comprehensive comparisons of our method against current SOTA approaches on Enc-VQA and InfoSeek benchmarks. 

\textbf{Zero-shot Setting.} Qwen2-VL-Oracle, which accesses ground-truth Wikipedia entries, establishes an upper-bound performance of 51.2\% on Enc-VQA and 57.9\% on InfoSeek. When using retrieved entries instead of gold references, our proposed CoRe-MMRAG achieves the best results among all methods, reaching 42.9\% accuracy on InfoSeek and 20.1\% on Enc-VQA, surpassing the second-best method, Qwen2-VL-1-Stage, by margins of 2.0\% and 2.2\%, respectively. Qwen2-VL-2-Stage yields suboptimal results due to potential information loss in unimodal reranking. Qwen2-VL-MMSTaR shows only slight improvements, indicating that current Chain-of-Thought reasoning remains limited for knowledge-intensive VQA. Nonetheless, it still outperforms the parametric-only baseline, highlighting the benefit of retrieved multimodal knowledge. Notably, in zero-shot settings, the performance gain on InfoSeek is more substantial than on Enc-VQA. This discrepancy is likely due to the larger and more complex knowledge base of Enc-VQA, which introduces greater retrieval noise and increases the difficulty of both PRKI and VTKI.

\textbf{Fine-tuned Setting.} Our method maintains superior performance, achieving 46.5\% on InfoSeek and 27.2\% on Enc-VQA. Furthermore, it exhibits the largest improvements comparing zero-shot setting, with gains of 3.6\% on InfoSeek and 7.1\% on Enc-VQA. Qwen2-VL-2-Stage, trained with $\mathcal{L}_{\mathrm{VTKI}}$ and $\mathcal{L}_{\mathrm{SFT}}$, shows an improvement of 2.3\% improvement over its zero-shot performance on InfoSeek and 6.1\% on Enc-VQA. Qwen2-VL-1-Stage, fine-tuned with $\mathcal{L}_{\mathrm{SFT}}$, achieves gains of 2.1\% on InfoSeek and 6.4\% on Enc-VQA. Qwen2-VL-MMSTaR demonstrates limited improvement, primarily due to the inadequate quality of its self-generated training instances. 

\begin{figure}[!htb]
  \vskip -0.11in
  \centering
  \includegraphics[width=0.89\columnwidth]{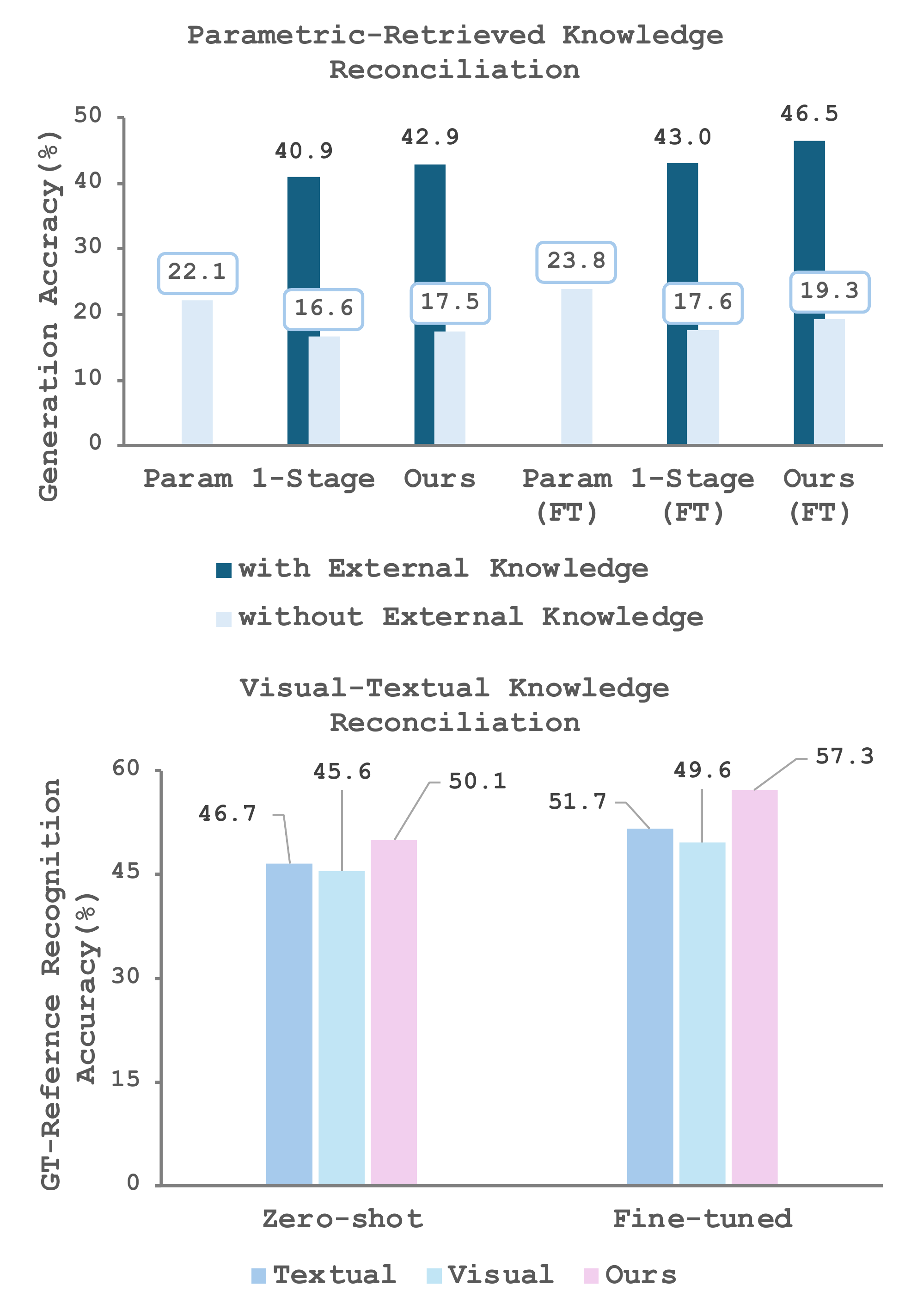}
  \vskip -0.09in
  \caption{Effectiveness of Our Method in Mitigating PRKI and VTKI. \textit{Top}: Evaluation of Parametric-Retrieved Knowledge Reconciliation, comparing our proposed method with Qwen-2-VL-Param (Param) and Qwen2-VL-1-Stage (1-Stage) under zero-shot and fine-tuned settings. \textit{Bottom}: Evaluation of Visual-Textual Knowledge Reconciliation, showing that our method improves ground-truth entry recognition through both textual and visual modalities across both settings.}
  \label{fig:conflicts_resolution}
  \vskip -0.1in
\end{figure}

\begin{table*}[!htbp]
    \centering
    \setlength{\tabcolsep}{.40em}
    \resizebox{0.85\linewidth}{!}{
    \begin{tabular}{lccclccclllllllllll}
    \toprule
    &  & \multicolumn{3}{c}{\textbf{Recall@1}} &  & \multicolumn{3}{c}{\textbf{Recall@2}} &  & \multicolumn{3}{c}{\textbf{Recall@5}} & \multicolumn{1}{c}{} & \multicolumn{3}{c}{\textbf{Recall@$k$ (K$>$5)}} &  & \textbf{All} \\ 
    \cline{3-5} \cline{7-9} \cline{11-13} \cline{15-17} \cline{19-19} 
    \textbf{Model} &  & U-Q & U-E & Acc &  & U-Q & U-E & Acc &  & \multicolumn{1}{c}{U-Q} & \multicolumn{1}{c}{U-E} & Acc &  & \multicolumn{1}{c}{U-Q} & \multicolumn{1}{c}{U-E} & Acc &  & Acc \\ 
    \midrule
    Param   &  & 25.0 & 25.3 & 25.2 &  & 24.4 & 23.9 & 24.1 &  & 20.7 & 22.3 & 21.8 &  & \textbf{19.3} & 15.7 & 16.4 &  & 22.1 \\
    \midrule
    1-Stage$\dagger$      & \\
    \hspace{0.4cm}Top-1 &  & \underline{59.1} & \underline{58.6} & \underline{58.8} &  & 30.1 & 27.3 & 28.0 &  & 22.0 & 22.4 & 22.1 &  & 15.4 & 15.6 & 15.6 &  & 39.6 \\
    \hspace{0.4cm}Top-2 &  & 58.4 & 57.9 & 58.2 &  & \underline{46.5} & \underline{45.7} & \underline{46.0} &  & 22.6 & 25.9 & 25.1 &  & 15.5 & 15.9 & 15.8 &  & 40.4 \\
    \hspace{0.4cm}Top-5 &  & 56.7 & 56.2 & 56.4 &  & 45.6 & 44.8 & 44.7 &  & \underline{32.3} & \underline{34.4} & \underline{33.8} &  & 16.8 & \textbf{16.5} & 16.6 &  & 40.9 \\
    \midrule
    Ours  \\
    \rowcolor{Mygreen!20}
    \hspace{0.4cm}Top-1 &  & \textbf{61.5} & \textbf{61.8} & \textbf{61.7} &  & 31.7 & 27.7 & 28.6 &  & 22.8 & 22.8 & 22.8 &  & 16.6 & 15.0 & 15.4 &  & 40.7 \\
    \rowcolor{Mygreen!20}
    \hspace{0.4cm}Top-2 &  & 61.0 & 61.2 & 61.2 &  & \textbf{48.4} & \textbf{49.0} & \textbf{48.9} &  & 23.8 & 27.3 & 26.4 &  & 16.9 & 15.7 & 16.0 &  & 42.9 \\
    \rowcolor{Mygreen!20}
    \hspace{0.4cm}Top-5 &  & 59.7 & 59.3 & 59.4 &  & 46.2 & 47.8 & 47.6 &  & \textbf{33.2} & \textbf{35.8} & \textbf{35.2} &  & 17.5 & 16.3 & \textbf{16.7} &  & \textbf{42.9} \\ 
    \bottomrule
    \end{tabular}
    }
    \caption{Performance (\%) with different numbers of retrieved entries. Recall@$k$ indicates the presence of ground-truth entry in top-$k$ retrieved results, with accuracy reported across Unseen-Questions (U-Q) and Unseen-Entities (U-E) settings.}
    \label{tab:Ablation test prompt}
    \vskip -0.2in
\end{table*}

\textbf{Knowledge Reconciliation.} Figure~\ref{fig:conflicts_resolution} illustrates the effectiveness of our method in addressing both VTKI and PRKI under zero-shot and fine-tuned settings. In the zero-shot setting, the model identifies 46.7\% of the ground truth Wikipedia entries using the textual modality and 45.6\% using the visual modality, with over 40\% of the entities correctly recognized by both. Our method improves this recognition rate to 50.1\% demonstrating the effectiveness of our method in handling VTKI. For PRKI, our method achieves better retention of parametric knowledge compared to the 1-Stage baseline, indicating more robust reconciliation of parametric-retrieved knowledge inconsistency. In the fine-tuned setting, the model’s ability to leverage multimodal knowledge for identifying ground-truth Wikipedia entries further improves, increasing from 50.1\% to 57.3\%, demonstrating the effectiveness of the objective $\mathcal{L}_{\mathrm{VTKI}}$. However, we observe that the 1-Stage baseline, trained solely with $\mathcal{L}_{\mathrm{SFT}}$, tends to compromise the retention of correct parametric knowledge, resulting in a larger performance drop (23.8\% to 17.6\%) relative to the original parametric outputs. In contrast, our method better preserves parametric knowledge, with a modest drop (23.8\% to 19.3\%), indicating the effectiveness of the $\mathcal{L}_{\mathrm{PRKI}}$ objective.

\subsection{Ablation Studies}
To validate the effectiveness of our approach, we conduct ablation studies on the InfoSeek validation set.

\textbf{Effect of Retrieved Entry Count.} Increasing the number of retrieved entries generally leads to improved overall recall accuracy. However, its impact varies across different Recall@$k$ groups, where Recall@$k$ indicates whether the ground-truth entry is included in the top-$k$ retrieved results. As shown in Table~\ref{tab:Ablation test prompt}, we observe a clear relationship between the number of retrieved entries (Top-$m$) and the Recall@$k$ performance. The first case is when $m \leq k$, meaning the number of retrieved entries does not exceed the evaluation threshold. In this setting, increasing $m$ directly expands the candidate pool, which generally leads to consistent performance improvements. For example, in the Recall@5 group, increasing $m$ from 2 to 5 improves recall accuracy, with the 1-Stage baseline rising from 25.1\% to 33.8\% and our method from 26.4\% to 35.2\%, as more relevant entries are included in the input. In contrast, when $m>k$, the inclusion of additional entries yields marginal or even negative returns, likely due to the introduction of irrelevant or noisy information. This effect is particularly evident in lower Recall@$k$ groups, where precision is more sensitive to input quality. For instance, in the Recall@2 group, increasing $m$ from 2 to 5 leads to a decrease in accuracy, from 46.0\% to 44.7\% for the 1-Stage baseline and from 48.9\% to 47.6\% for our method. 

Moreover, the impact of $m$ varies across evaluation scenarios. While increasing the number of external entries initially benefits both Unseen-Q and Unseen-E, the latter demonstrates more stable and robust improvements. In contrast, performance degradation caused by excessive retrieval is more pronounced in Unseen-Q, highlighting its greater sensitivity to noisy or irrelevant knowledge.

\textbf{Effect of Different Prompt.}
Table~\ref{tab:Ablation test prompt} presents a comprehensive zero-shot performance analysis between Qwen2-VL-1-Stage and our proposed method. For PRKI resolution, CoRe-MMRAG exhibits enhanced robustness to knowledge noise through carefully designed prompts. In Recall@1 group, when increasing retrieved entries from 1 to 2, our method maintains stable performance with accuracy shifting from 61.7\% to 61.2\%, while Qwen2-VL-1-Stage shows larger degradation from 58.8\% to 58.2\%. This advantage becomes more evident in the setting of Unseen-Questions, where our method preserves accuracy from 61.5\% to 61.0\% compared to the Qwen2-VL-1-Stage's significant drop from 59.1\% to 58.4\%.

Moreover, our approach demonstrates superior multimodal knowledge integration capabilities. In Recall@2 group, increasing retrieved entries from 1 to 2 yields substantially larger gains as our method improves from 28.6\% to 48.9\% versus Qwen2-VL-1-Stage advancing from 28.0\% to 46.0\%. The improvement margin widens further in Unseen-Entities scenarios, with our method achieving progress from 27.7\% to 49.0\% while the baseline moves from 27.3\% to 45.7\%, confirming the effectiveness of our method on visual-textual fusion, as visualized in Figure~\ref{fig:conflicts_resolution}.

\begin{table}[t]
    \begin{center}
    \begin{small}
    \begin{tabular}{lccc}
    \toprule
    \multirow{2}{*}{\textbf{Methods}} & \multicolumn{3}{c}{\textbf{Recall@5 (\%)}} \\
    \cmidrule(r){2-4}
     & Unseen-Q & Unseen-E & All  \\
    \midrule
    \textbf{Ours} & 45.2& 46.9& 46.5\\
    \hspace{0.4cm}w/o $\mathcal{L}_{\mathrm{PRKI}}$ & 44.1 & 46.0 & 45.5  \\
    \hspace{0.4cm}w/o $\mathcal{L}_{\mathrm{VTKI}}$ & 44.2& 45.6& 45.3  \\
    \hspace{0.4cm}w/o $\mathcal{L}_{\mathrm{SFT}}$ & 43.3 & 45.0 & 44.6  \\
    \bottomrule
    \end{tabular}
    \end{small}
    \end{center}
    \caption{Ablation study on training objectives.}
    \label{tab:encoder_comparison}
    \vskip -0.2in
\end{table}

\textbf{Effect of Training Objectives.} Table \ref{tab:encoder_comparison} demonstrates the contribution of each training objective. Removing any objective leads to performance degradation, with $\mathcal{L}_{\mathrm{SFT}}$ causing the most significant decline at 1.9\% in overall performance. The absence of $\mathcal{L}_{\mathrm{PRKI}}$ and $\mathcal{L}_{\mathrm{VTKI}}$ also leads to notable performance drops at 1.0\% and 1.2\% respectively, indicating both objectives are essential for effective knowledge conflict and inconsistency resolution. Our full model achieves the best performance across all metrics, validating the complementary nature of these objectives.

\section{Conclusion}
In this paper, we present CoRe-MMRAG, a novel framework that addresses two critical challenges in MMRAG: parametric-retrieved knowledge inconsistency and visual-textual knowledge inconsistency. CoRe-MMRAG follows a four-stage pipeline that effectively reconciles internal parametric knowledge with externally retrieved information and leverages joint similarity assessment to integrate complementary visual and textual signals. To further enhance its capabilities, we introduced a specialized training paradigm with three targeted objectives focused on knowledge source selection, multimodal integration, and answer generation. Extensive experiments on InfoSeek and Enc-VQA benchmarks demonstrate the effectiveness of our approach, achieving performance gains of 5.6\% and 9.3\% over baseline methods.

\section{Limitations and Future Works}
Despite the promising results, our work has several important limitations. First, our framework's effectiveness is heavily dependent on the initial retrieval quality using Eva-CLIP-8B. As demonstrated in our experiments, the retrieval performance (Recall@1) remains relatively low at 45.6\% for InfoSeek and 13.3\% for Enc-VQA. This limited retrieval performance creates a ceiling effect for the overall system performance, suggesting that improvements in the initial retrieval stage could lead to significant gains in the final results. Second, our approach faces substantial computational challenges. The four-stage process and multiple training objectives require significant computational resources, with training taking approximately 10 hours on 8 H100 GPUs. The model requires substantial memory to process both visual and textual knowledge simultaneously, which may limit its practical deployment in resource-constrained environments. Finally, there are limitations in terms of dataset coverage and generalization. While we demonstrate strong performance on two specific KB-VQA benchmarks, the effectiveness of our approach on other types of multimodal tasks or real-world scenarios remains unexplored. The model's performance on rare entities or edge cases in the knowledge base may be suboptimal, and there exists a potential domain gap between the Wikipedia knowledge base used for training and real-world applications. Future work should address these limitations to improve the practical applicability of our approach.

\clearpage
\bibliography{custom}

\clearpage
\appendix
\section{Appendix}
\subsection{Prompts}
In this section, we present the prompts used in different MMRAG pipelines, including Qwen2-VL-Param, Qwen2-VL-Oracle, Qwen2-VL-1-Stage, Qwen2-VL-2-Stage, Qwen2-VL-MMSTaR, and our proposed CoRe-MMRAG. Each prompt is illustrated with the format, an example, and the corresponding final answer.

\begin{small}
\begin{tcolorbox}[promptbox, title= Prompt for Qwen2-VL-Param]
\textbf{Prompt format:}

\begin{lstlisting}[language=Python]
Question: <image>
{question}
Please use parametric knowledge answer the question within 5 words
\end{lstlisting}

\textbf{Example:}

\begin{lstlisting}
Question: <image>
Which historic county does this village belong to? 
Please use parametric knowledge answer the question within 5 words
\end{lstlisting}

\textbf{Answer:} 
\begin{lstlisting}
"Scotland"
\end{lstlisting}

\end{tcolorbox}    
\end{small}

The prompt for Qwen2-VL-Oracle is constructed using the ground-truth Wikipedia entry. Specifically, \lstinline{wiki_title_gt} and \lstinline{wiki_content_gt} refer to the title and textual content of the ground-truth entry, respectively.

\begin{small}
\begin{tcolorbox}[promptbox, title= Prompt for Qwen2-VL-Oracle]
\textbf{Prompt format:}

\begin{lstlisting}[language=Python]
Based on the retrieved document, answer the question
<image>
{question} within 5 words
Context:
Reference Image:<image>
Wiki title: {wiki_title_gt}
Wiki content:{wiki_content_gt}
\end{lstlisting}

\textbf{Example:}

\begin{lstlisting}[language=Python]
Based on the retrieved document, answer the question
<image>
What is the parent organization of this building? within 5 words.
Context:
Reference Image:<image#gt>
Wiki title: Colonial Williamsburg
Wiki content:Colonial Williamsburg is a living-history museum and private foundation presenting a part of the historic diseconstructions...
\end{lstlisting}

\textbf{Answer:} 
\begin{lstlisting}
"The Colonial Williamsburg Foundation"
\end{lstlisting}

\end{tcolorbox}    
\end{small}

The following is the prompt for Qwen2-VL-1-Stage, where \lstinline{wiki_title_A/B/C/D/E} and \lstinline{wiki_content_A/B/C/D/E} represent the titles and textual contents of the top-5 retrieved Wikipedia entries.

\begin{small}
\begin{tcolorbox}[promptbox, title= Prompt for Qwen2-VL-1-Stage]
\textbf{Prompt format:}

\begin{lstlisting}[language=Python]
Based on the retrieved document, answer the question
<image>
{question} within 5 words
Reference A:
<image#A>
Wiki title: {wiki_title_A}
Wiki content:{wiki_content_A}
Reference B:
<image#B>
Wiki title: {wiki_title_B}
Wiki content:{wiki_content_B}
Reference C:
<image#C>
Wiki title: {wiki_title_C}
Wiki content:{wiki_content_C}
Reference D:
<image#D>
Wiki title: {wiki_title_D}
Wiki content:{wiki_content_D}
Reference E:
<image#E>
Wiki title: {wiki_title_E}
Wiki content:{wiki_content_E}
\end{lstlisting}

\textbf{Example:}

\begin{lstlisting}[language=Python]
Based on the retrieved document, answer the question
<image>
What is the width (in kilometre) of this lake? within 5 words.

Reference A 
<image#A>
Wiki title: Wadi Numeira
Wiki content: Wadi Numeira is a Wadi in Jordan that is known for its deep gorge cut through the sandstone...

Reference B
<image#B>
Wiki title: Kalya
Wiki content: Kalya is an Israeli settlement organized as a kibbutz in the West Bank..

Reference C 
<image#C>
Wiki title: Ein Gedi
Wiki content: Ein Gedi, also spelled En Gedi, meaning \"spring of the kid\",is an oasis and a nature reserve in Israel...

Reference D 
<image#D>
Wiki title: Dead Sea
Wiki content: The Dead Sea, also known by other names, is a salt lake bordered by Jordan to the east and Israel...

Reference E 
<image#E>
Wiki title: Jewish National Fund
Wiki content: Jewish National Fund was founded in 1901 to buy and develop land in Ottoman Syria..."
\end{lstlisting}

\textbf{Answer:} 
\begin{lstlisting}
"10"
\end{lstlisting}

\end{tcolorbox}    
\end{small}

The following is the prompt for Qwen2-VL-2-Stage, where \lstinline{wiki_title_A/B/C/D/E} and \lstinline{wiki_content_A/B/C/D/E} denote the titles and textual contents of the top-5 retrieved Wikipedia entries. \lstinline{wiki_title_select} and \lstinline{wiki_content_select} refer to the title and content of the entry selected during the first-stage reranking.

\begin{small}
\begin{tcolorbox}[promptbox, title= Prompt for Qwen2-VL-2-Stage]
\textbf{Prompt format for rerank:}

\begin{lstlisting}[language=Python]
Identify the most similar wiki context to the question
<image>
{question}
Context A:
Wiki title: {wiki_title_A}
Wiki content:{wiki_content_A}
Context B:
Wiki title: {wiki_title_B}
Wiki content:{wiki_content_B}
Context C:
Wiki title: {wiki_title_C}
Wiki content:{wiki_content_C}
Context D:
Wiki title: {wiki_title_D}
Wiki content:{wiki_content_D}
Context E:
Wiki title: {wiki_title_E}
Wiki content:{wiki_content_E}
The answer shold be provided in format: [Reference X] where X is the most similar reference (A/B/C/D/E)
\end{lstlisting}

\textbf{Prompt format for generation:}
\begin{lstlisting}
Based on the retrieved document, answer the question 
<image> 
{question} within 5 words.
Context:
Wiki title: {wiki_title_select}
Wiki content: {wiki_content_select}
\end{lstlisting}

\textbf{Example:}

\begin{lstlisting}[language=Python]
# Rerank:
...Context A: ...
# See Prompt for Qwen2-VL-1-Stage

The answer shold be provided in format: [Reference X] where X is the most similar reference (A/B/C/D/E)

# Generation:
Based on the retrieved document, answer the question 
<image> 
What is the width (in kilometre) of this lake? within 5 words

Context: 
Wiki title: Jewish National Fund
Wiki content: Jewish National Fund was founded in 1901 to buy and develop land in Ottoman Syria...
\end{lstlisting}

\textbf{Answer:} 
\begin{lstlisting}
"Reference E"
"Not given"
\end{lstlisting}

\end{tcolorbox}    
\end{small}

The following is the prompt for Qwen2-VL-MMSTaR, where \lstinline{wiki_title_A/B/C/D/E} and \lstinline{wiki_content_A/B/C/D/E} represent the titles and textual contents of the top-5 retrieved Wikipedia entries.

\begin{small}
\begin{tcolorbox}[promptbox, title= Prompt for Qwen2-VL-MMSTaR (Part I)]
\textbf{Prompt format:}

\begin{lstlisting}[language=Python]
Based on the retrieved document, answer the question
<image>
{question} within 5 words
Reference A:
<image#A>
Wiki title: {wiki_title_A}
Wiki content:{wiki_content_A}
\end{lstlisting}


\end{tcolorbox}    
\end{small}

\begin{small}
\begin{tcolorbox}[promptbox, title= Prompt for Qwen2-VL-MMSTaR (Part II)]
\textbf{Prompt format:}

\begin{lstlisting}[language=Python]
Reference B:
<image#B>
Wiki title: {wiki_title_B}
Wiki content:{wiki_content_B}
Reference C:
<image#C>
Wiki title: {wiki_title_C}
Wiki content:{wiki_content_C}
Reference D:
<image#D>
Wiki title: {wiki_title_D}
Wiki content:{wiki_content_D}
Reference E:
<image#E>
Wiki title: {wiki_title_E}
Wiki content:{wiki_content_E}
Note: Analyze step-by-step and provide final answer in format: []
\end{lstlisting}

\textbf{Example:}

\begin{lstlisting}[language=Python]
Based on the retrieved document, answer the question
<image>
What place inflows lake? within 5 words.
Reference A: ...
# See Prompt for Qwen2-VL-1-Stage 
Note: Analyze step-by-step and provide final answer in format: []
\end{lstlisting}

\textbf{Answer:} 
\begin{lstlisting}
"Step 1: Key Information Analysis
Query Image Analysis:
The query image shows a scenic view of a lake surrounded by mountains and a town. The lake is prominent in the center, with a town on its shores.

Question Analysis:
The question asks about the place that inflows into the lake.


Step 2: Reference Comparison:
Reference A: Describes Gsteigwiler, a village near a lake.

Reference B: Describes Walensee, a lake in Switzerland.

Reference C: Describes Schnepfau, a village near a lake.

Reference D: Describes Lake Lucerne, a lake in Switzerland.

Reference E: Describes Lake Brienz, a lake in Switzerland.

Relevant Reference Selection:
Reference B (Walensee) and Reference D (Lake Lucerne) both describe lakes in Switzerland, but Reference B is more likely to have a river inflowing into it, as it is a larger lake.


Step 3: Answer Generation
Answer:The place that inflows into Lake Walensee is the Seez river. According to the description in Reference B, the Seez, Murgbach, and Linth rivers lead to the lake. The Seez river is the most likely candidate for inflowing into the lake, so the final answer is [Seez river]"
\end{lstlisting}

\end{tcolorbox}    
\end{small}

The following is the prompt for CoRe-MMRAG, where \lstinline{wiki_title_A/B/C/D/E} and \lstinline{wiki_content_A/B/C/D/E} represent the titles and textual contents of the top-5 retrieved Wikipedia entries.

\begin{small}
\begin{tcolorbox}[promptbox, title= Prompt for CoRe-MMRAG]
\textbf{Prompt format:}

\begin{lstlisting}[language=Python]
Question: 
<image>
{question}
Step 1:
If the question with image is directly answerable, provide the answer in [] less than 5 words.
If not, proceed to Step 2.

Step 2: 
Identify the most similar Wikipedia reference to the {question}
Reference A:
<image#A>
Wiki title: {wiki_title_A}
Wiki content:{wiki_content_A}
Reference B:
<image#B>
Wiki title: {wiki_title_B}
Wiki content:{wiki_content_B}
Reference C:
<image#C>
Wiki title: {wiki_title_C}
Wiki content:{wiki_content_C}
Reference D:
<image#D>
Wiki title: {wiki_title_D}
Wiki content:{wiki_content_D}
Reference E:
<image#E>
Wiki title: {wiki_title_E}
Wiki content:{wiki_content_E}

Step 3:
Based on the most similar reference, answer the question 
<image>
{question} 
within 5 words

Step 4:
If the answers from Step 1 and Step 3 differ, determine which one is the final answer, output the final answer in less than 5 words.
\end{lstlisting}

\textbf{Example:}

\begin{lstlisting}[language=Python]
Question:
<image> 
What is the width (in kilometre) of this lake?
Step 1:
If the question with image is directly answerable, provide the answer in [] less than 5 words.
If not, proceed to Step 2.

Step 2: 
Identify the most similar Wikipedia reference to What is the width (in kilometre) of this lake?
Reference A...
# See Prompt for Qwen2-VL-1-Stage

Step 3:
Based on the most similar reference, answer the question 
<image>
{question} 
within 5 words

Step 4:
If the answers from Step 1 and Step 3 differ, determine which one is the final answer, output the final answer in less than 5 words.
\end{lstlisting}

\textbf{Answer:} 
\begin{lstlisting}
"2.8 km"
\end{lstlisting}

\end{tcolorbox}    
\end{small}



\subsection{Case Study}
\subsubsection{VTKI Examples}
The cases for VTKI are illustrated in Figure~\ref{fig:case_1}. For each input query, we present the top-5 retrieved references (A–E) ranked by embedding similarity between the reference images and the query image, computed using EVA-CLIP-8B. The examples in the figure demonstrate that the Qwen-2-VL model produces different outputs when ranking is based solely on image modality versus text modality. When relying only on image similarity, the model may incorrectly select visually similar but semantically irrelevant references. For instance, in the second row of Figure~\ref{fig:case_1}, the model selects reference C as the most similar, but its associated text discusses Bursaphelenchus xylophilus, which is unrelated to the question "How many offspring can this bird produce at the same time?"
Conversely, when using only textual information, as shown in the fourth row of Figure~\ref{fig:case_1}, the model mistakenly selects reference B over reference C. Both references describe food items resembling pudding, but without visual cues, the model cannot determine which is more appropriate. These examples highlight the VTKI problem, where unimodal approaches may lead to incorrect reference selection. In contrast, our proposed CoRe-MMRAG method effectively addresses this issue by leveraging both modalities. When unimodal outputs diverge but include the correct answer, CoRe-MMRAG enables the model to identify and choose the correct reference.

\subsubsection{PRKI Examples}
\begin{figure*}[ht]
    \begin{center}
    \centerline{\includegraphics[width=1.0\textwidth]{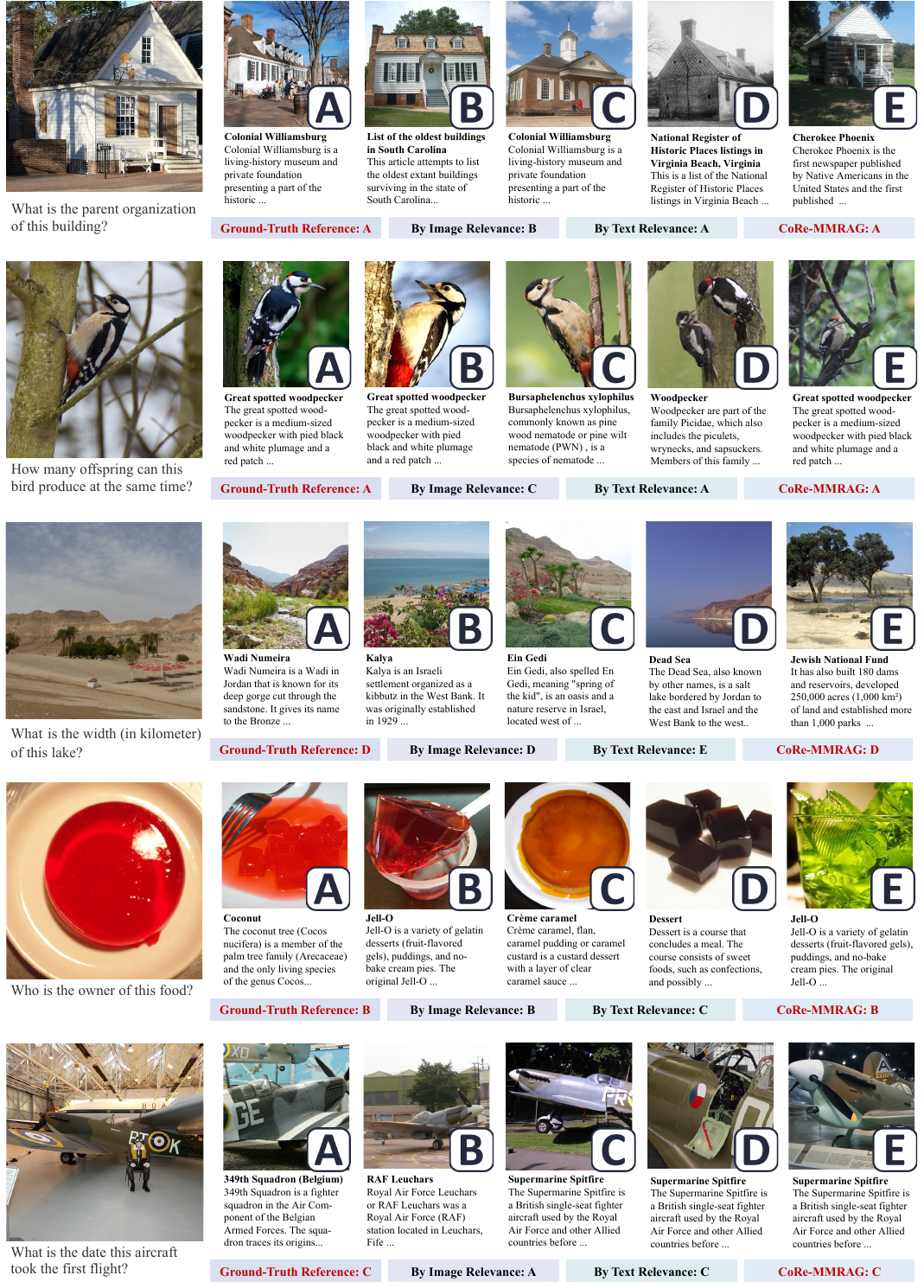}}
    \caption{Qualitative results on sample image-question pairs from the InfoSeek dataset. The leftmost column displays the input query image. References A–E are ordered by descending embedding similarity (computed using EVA-CLIP-8B) between the reference images and the query image. "By image relevance" presents the Qwen-2-VL output of most relevant references based solely on reference images, while "By text relevance" relies only on reference texts. Our proposed CoRe-MMRAG model leverages multimodal information to select the most relevant references.}
    \label{fig:case_1}
    \end{center}
\vskip -0.4in
\end{figure*}

\begin{figure*}[ht]
    \begin{center}
    \centerline{\includegraphics[width=1.0\textwidth]{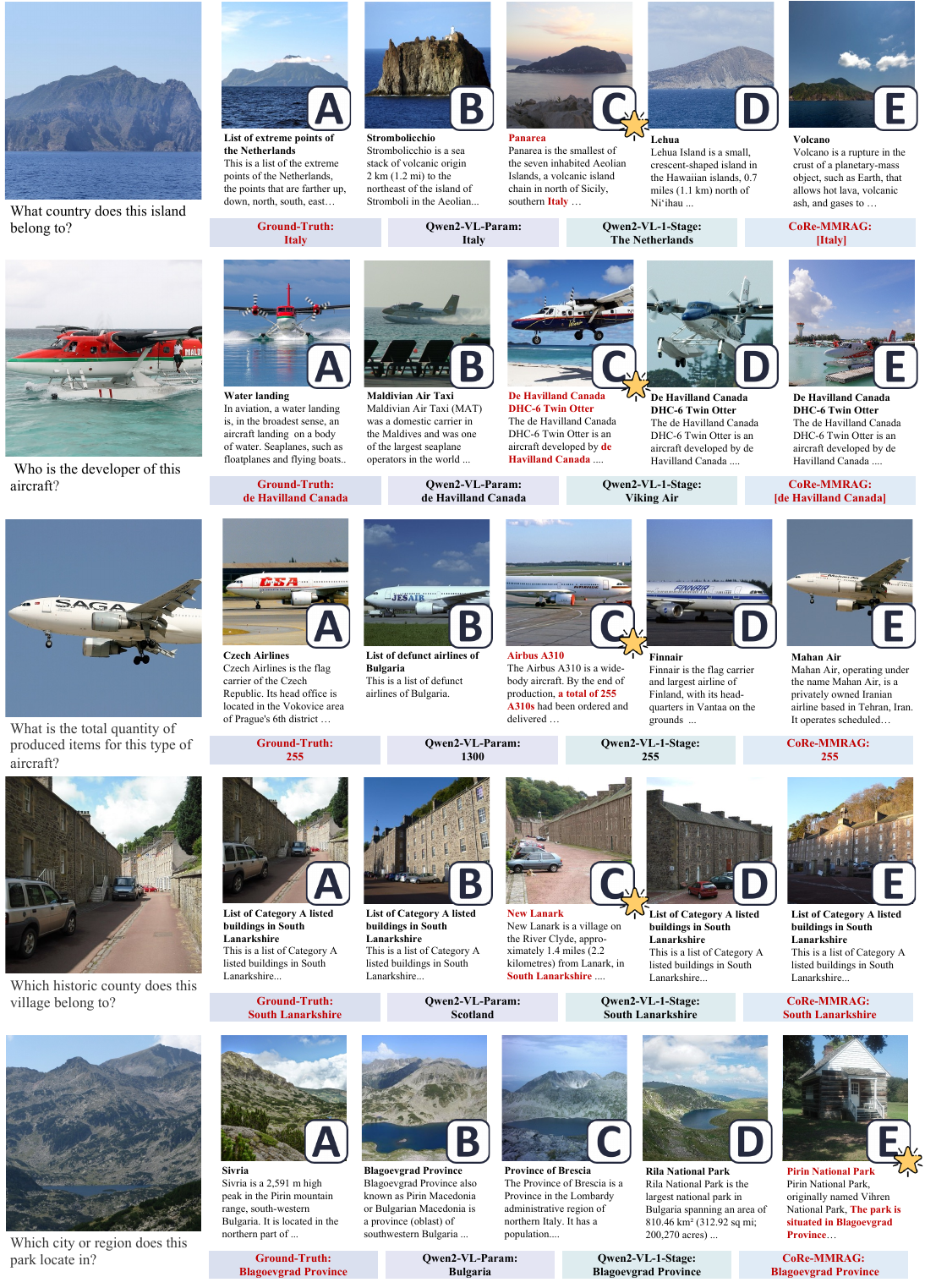}}
    \caption{Qualitative final answer outputs on sample image-question pairs from the InfoSeek dataset. The leftmost column shows the input query image. References A–E are ordered by descending embedding similarity (computed using EVA-CLIP-8B) between the reference images and the query image. A star symbol indicates the ground-truth reference.}
    \label{fig:case_2}
    \end{center}
\vskip -0.4in
\end{figure*}

The cases for PRKI are illustrated in Figure~\ref{fig:case_2}, which follows the same ranking settings as Figure~\ref{fig:case_1}. These examples demonstrate that the presence of noisy information in external data can negatively impact the model's ability to express accurate parametric knowledge, leading to incorrect outputs. In such PRKI scenarios, our proposed model effectively mitigates this issue. Notably, the first and second rows in Figure~\ref{fig:case_2} show that our method successfully outputs the correct parametric knowledge, explicitly marked with square brackets ("[]"). 

\end{document}